# Automatic Meter Classification of Kurdish Poems

**Aso Mahmudi**
Faculty of New Sciences and Technologies,
University of Tehran, Iran
aso.mahmudi@ut.ac.ir

**Hadi Veisi**
Faculty of New Sciences and Technologies,
University of Tehran, Iran
h.veisi@ut.ac.ir

**Abstract**

Most of the classic texts in Kurdish literature are poems. Knowing the meter of the poems is helpful for correct reading, a better understanding of the meaning, and avoidance of ambiguity. This paper presents a rule-based method for automatic classification of the poem meter for the Central Kurdish language. The metrical system of Kurdish poetry is divided into three classes of quantitative, syllabic, and free verses. As the vowel length is not phonemic in the language, there are uncertainties in syllable weight and meter identification. The proposed method generates all the possible situations and then, by considering all lines of the input poem and the common meter patterns of Kurdish poetry, identifies the most probable meter type and pattern of the input poem. Evaluation of the method on a dataset from VejinBooks Kurdish corpus resulted in 97.3% of precision in meter type and 96.2% of precision in pattern identification.

**Keywords:** Computational Linguistics, Poetry classification, Central Kurdish

## 1 Introduction

Kurdish is an Indo-Iranian language spoken by millions of peoples mainly in Iraq, Turkey, Iran, and Syria. Among various dialects of Kurdish, in this paper, we focus on the standard literary form of the central dialect (Standard Central Kurdish, SCK), which has produced literary texts more than other dialects in the past century (Hassanpour, 1992). A greater part of classical Kurdish literature is in poetry form, maybe because poems are easier to memorize when powerful neighboring languages, i.e. Arabic, Persian and Turkish, confined its writing in the past centuries.

The identification of form (rhyming scheme) and meter (rhythmic structure) of poetry is a time-consuming and complicated task for beginners. An automatic application can help students and inexpert poets to learn and correct their mistakes. There are three kinds of Kurdish poem meter: Quantitative (syllable weight rhythm), Syllabic (syllable count rhythm), and Free verses. In this paper, we introduce a rule-based method for the automatic classification of Kurdish poem meter. The method, first, decomposes each line of the poem into its syllables. Then, it identifies the repetition pattern in syllable weight and number. But, the poetry related tasks makes more significant challenges for natural language processing than any other genre (Kesarwani, 2018). In this task, there are issues too. In the writing system of Kurdish, correspondence between some graphemes and phonemes is not one-to-one; therefore, the syllabification process confronts ambiguities. In addition, syllable weight is not a distinctive concept in Kurdish, and it is probable to change the weight of some syllables in a poem without altering the meaning.

The proposed method, utilizes a rule-based method of SCK grapheme-to-phonemes convertor (Mahmudi & Veisi, 2021), which syllabifies the input poem. Then, for the analysis and detection of the poem meter, we consider all possible patterns of each line. Eventually, we analyze the whole poem to calculate a score for each common quantitative pattern. If a pattern repeats in all lines, the proposed method classifies it as quantitative. Else, if most of the lines have equal syllable count, the poem is a syllabic verse; otherwise, it is a free verse.

The rest of the paper is organized as follows: Section 2 reviews phonology and the alphabet of Standard Central Kurdish and the common types of meters of Kurdish poems. Section 3 presents the steps of the proposed method for the classification of Kurdish poems. Section 4 describes the test dataset and results. Section 5 gives conclusions and further works.

## 2 Background and Related Works

### 2-1 Phonemes, Alphabet, and Syllables of SCK

There are 37 phonemes in SCK, including 8 vowels and 29 consonants (Mahmudi & Veisi, 2021). This study uses the Hawar alphabet (standard Latin script for Northern Kurdish) with changes in some consonants. Table 1 compares IPA and the Standard Arabic alphabet of the Kurdish with this study's transcription of consonants.

Table 1: Consonants of Standard Central Kurdish

| *Kurdish Alphabet* | ئ | ب | پ | ت | ج | چ | ح | خ | د | ر | ڕ | ز | ژ | س | ش | ع | غ | ف | ڤ | ق | ک | گ | ل | ڵ | م | ن | و | ھ | ی |
|---|---|---|---|---|---|---|---|---|---|---|---|---|---|---|---|---|---|---|---|---|---|---|---|---|---|---|---|---|---|
| *IPA* | ʔ | b | p | t | dʒ | tʃ | ħ | x | d | ɾ | r | z | ʒ | s | ʃ | ʕ | ɣ | f | v | q | k | g | l | ɫ | m | n | w | h | j |
| *This Study* | *ʔ* | *b* | *p* | *t* | *c* | *ç* | *ḧ* | *x* | *d* | *r* | *ř* | *z* | *j* | *s* | *ş* | *ɛ* | *ẍ* | *f* | *v* | *q* | *k* | *g* | *l* | *ł* | *m* | *n* | *w* | *h* | *y* |

As the syllable weight is the essential material in the identification of poem meter, we will discuss SCK vowel's length more precisely. are shorter in final unstressed positions, and the short vowel */e/* in word-final positions can be pronounced longer (McCarus, 1958). The vowel */i/* is unstable in most environments and does not have a grapheme in the standard Kurdish alphabet (MacKenzie, 1961).

Table 2 describes the details of Standard Central Kurdish vowels. The long vowels *(/î, ê, a, o, û/)* are shorter in final unstressed positions, and the short vowel */e/* in word-final positions can be pronounced longer (McCarus, 1958). The vowel */i/* is unstable in most environments and does not have a grapheme in the standard Kurdish alphabet (MacKenzie, 1961).

Table 2: Vowels of Standard Central Kurdish

| **IPA** | **this study** | **Kurdish alphabet** | **description** | **normal length** |
|---|---|---|---|---|
| i | *î* | ی | close front unrounded | long |
| ɛ | *ê* | ێ | mid-open front unrounded | long |
| ä | *a* | ا | open front-central unrounded | long |
| o̞ | *o* | ۆ | mid back rounded | long |
| u | *û* | وو | close back rounded | long |
| ʊ | *u* | و | half-close back-central rounded | short |
| a | *e* | ە | open front unrounded | short |
| ɪ | *i* | | half-close front-central unrounded | short |

The Kurdish alphabet, which is adapted from the Perso-Arabic script, has ambiguities in three cases (Mahmudi & Veisi, 2021):

1. The letter "ی" indicates both consonant */y/* and vowel */î/*.
2. The letter "و" can represent the consonant */w/* and the short vowel */u/*. In addition, if this letter comes up twice ("وو"), it can be long vowel */û/* or combinations */uw/*, */uw/* or */ww/*.
3. There is no letter for short vowel */i/* in Arabic script of Kurdish.

In the syllable structure of Kurdish, the nucleus is always a vowel, and the onset is one or two consonants. In two-consonant onsets, the second consonant must be */w/* or */y/* and the first consonant must be stop, fricative, or affricate. Coda has zero up to three consonants. Three-consonant coda is rare, occurs only in some dialects (Karimi-Doostan, 2002), and was not observed in our dataset of Kurdish poetry. The sonority value decreases in the coda for obeying the sonority sequencing principle (Zahedi et al., 2012). Table 3 presents syllable types and their normal weight in SCK.

Table 3: Syllable Types in Standard Central Kurdish

| No. | Syllable Type | example | normal weight |
|---|---|---|---|
| 1 | CV | */be/* ('with')<br>*/ba/* ('wind') | light<br>heavy |
| 2 | CVC | */ber/* ('front') | heavy |
| 3 | CVCC | */berd/* ('stone') | heavy |
| 4 | CVcCC | */řoyşt/* ('went') | heavy |
| 5 | CcV | */xwê/* ('salt') | heavy |
| 6 | CcVC | */xwên/* ('blood') | heavy |
| 7 | CcVCC | */xwênd/* ('read') | heavy |

Note: c=approximant, C=other consonants, V=Vowel.

## 2-2 Types of Kurdish Poems

As mentioned, the Kurdish language is a collection of dialects whose speakers live in Iran, Iraq, Turkey, Syria, and parts of Caucasus. Neighboring different nations has led Kurdish literature to enjoy the characteristics of different literature styles.

In the classical literature of Kurdish, there are three categories of poetic works: "quantitative (Arudi) verse", "beit (syllabic songs)" and "gorani (lyric songs)" (Mokri, 1950). Kurdish quantitative verses are an imitation of Arabic and especially Persian poetry (Jobraili, 2015), and in terms of meter, it is based on syllable weight, i.e., all lines of a poem have an equal number of syllables, repeating a pattern of light and heavy syllables. Beit and gorani have "syllabic meter". The syllabic or numerical meter is rooted in the ancient tradition of Iranian languages, and it has long existed among different ethnic groups in Iran (Mokri, 1950). There is evidence of a syllabic meter in pre-Islamic literature in the texts of the Zoroastrian and Manichaean rituals (Emmerick & Macuch, 2008). In a syllabic meter, the weight of the syllables and the place of stress do not affect the meter and only the total number of syllables in each line is important.

Fixed form poems in Kurdish are consist of lines that have an equal number of syllables. In most of the forms, like ghazal and mathnawi, even lines are rhyming, however, in some forms like mukhammas, rhyming is different. In modern literature of Kurdish "free verse" is a new style which is not limited to a fixed form and the number of syllables in each line may be different (Ghaderi, 2016). This study considers three types for Kurdish poems: Quantitative, Syllabic, and Free verses.

### 2-2-1 Syllabic Verses

In syllabic or numerical verses, only the number of syllables in feet is considered, and the syllable weight sequence is not following a specific pattern. Kurdish folk poems are syllabic verses (Bakir, 2004). There are three types of three, four, and five-syllable feet in Kurdish syllabic verses, which are repeated uniformly or alternately at each line (Gardi, 2014). Table 4 shows the types of syllabic verses in Kurdish and the way in which feet are combined to form each line. The most common type of syllabic verses in Kurdish is 10-syllabic.

Table 4: Types of syllabic verses in Kurdish poetry
(Frequencies from VejinBooks corpus, up to 2019/12/1)

| **type** | **feet order** | **frequency** |
|---|---|---|
| *5-syllabic* | =5 | 0 |
| *6-syllabic* | =3+3 | 0 |
| *7-syllabic* | =4+3 | 34 |
| *8-syllabic* | =4+4 | 159 |
| *9-syllabic* | =3+3+3 | 0 |
| *10-syllabic* | =5+5 | 2,020 |
| *11-syllabic* | =4+4+3 | 60 |
| *12-syllabic* | =4+4+4 or =3+3+3+3 | 14 |
| *13-syllabic* | =4+4+5 | 23 |
| *14-syllabic* | =4+3+4+3 | 31 |
| *15-syllabic* | =5+5+5 or =4+4+4+3 | 17 |
| *16-syllabic* | =4+4+4+4 | 19 |

**2-2-2 Quantitative Verses**

Quantitative meter is an arrangement of heavy (¯) and light (˘) syllables in a line of the poem as is found in Greek and Latin poems (Hayes, 2009). This type of meter fits for languages like Arabic, where the vowel length is distinctive and changes the meaning. Arabic has three short vowels [a i u] with distinctive long pairs [aː iː uː]. For example, in the following Arabic hemistich by Hafez (1325-1390), all vowels are pronounced with their normal lengths:

«الا يا ايُّهَا الساقي أدر كاساً و ناولها»

syllables: *ʔa laː yaː ʔay yu has saː qiː ʔa dir kaʔ san wa naː wil haː*
˘ ¯ ¯ ¯ ˘ ¯ ¯ ¯ ˘ ¯ ¯ ¯ ˘ ¯ ¯ ¯

However, in languages such as Persian and Kurdish, whose vowel length is not distinctive, to follow the metrical pattern of the poem, some syllables can be pronounced contrary to their natural weight. For example, in another hemistich of that poem which is in Persian, short vowel */e/* in the word */ha.me/* 'all', should be pronounced as */ha.meː/*, for preserving the meter:

«همه کارم ز خودکامی به بدنامی کشید آخر»

syllables: *hæ meː kɒː ræm ze xod kɒː miː be bæd nɒː miː ke šiː dɒː xer*
˘ ¯ ¯ ¯ ˘ ¯ ¯ ¯ ˘ ¯ ¯ ¯ ˘ ¯ ¯ ¯

In Kurdish, as shown in Table 3, only syllable which has a single consonant onset and short vowel nucleus and no coda (e.g. */be/*) are light and all other types of syllables are heavy. In Kurdish quantitative verses, such as the following line by Qani' (1898-1965), syllable weights are regularly pronounced as their natural weights:

«لەباتی من بلّێن بولبول نەخوێنێ قەت بە مل گوڵدا»

syllables: *le ba tî min bi łên bul bul ne xwê nê qet be mil guł da*
˘ ¯ ¯ ¯ ˘ ¯ ¯ ¯ ˘ ¯ ¯ ¯ ˘ ¯ ¯ ¯

However, for saving the meter, some syllables (4, 5, and 11) are pronounced differently, as in the following line by Piramerd (1867-1950):

«چەند ساڵ گوڵی هیوای ئێمە پێ پەست بوو تاکو پار»

| | 1 | 2 | 3 | 4 | 5 | 6 | 7 | 8 | 9 | 10 | 11 | 12 | 13 | 14 |
|---|---|---|---|---|---|---|---|---|---|---|---|---|---|---|
| syllables: | *çend* | *sał* | *gu* | *łî* | *hî* | *way* | *ʔê* | *me* | *pê* | *pest* | *bû* | *ta* | *ku* | *par* |
| normal weights: | ¯ | ¯ | ˘ | ˘ | ¯ | ¯ | ¯ | ˘ | ¯ | ¯ | ¯ | ¯ | ˘ | ¯ |
| meter pattern: | ¯ | ¯ | ˘ | ¯ | ˘ | ¯ | ˘ | ˘ | ¯ | ¯ | ˘ | ¯ | ˘ | ¯ |

Table 5 shows the most common patterns of Kurdish quantitative verses extracted from VejinBooks corpus (Vejinbooks Contributors, 2019).

Additionally, Aziz Gardi (2004) has conducted a comprehensive statistical study on quantitative verses of 82 Kurdish poets. Future works will benefit from its information.

Table 5 : Common patterns of Kurdish quantitative verses (VejinBooks corpus, up to 2019/12/1)

| rank | pattern title | syllable weight pattern | freq. | % |
|---|---|---|---|---|
| *1* | فاعلاتن فاعلاتن فاعلاتن فاعلن | –˘––/–˘––/–˘––/–˘– | 1044 | 27.14 |
| *2* | مفاعیلن مفاعیلن مفاعیلن مفاعیلن | ˘–––/˘–––/˘–––/˘––– | 999 | 25.97 |
| *3* | مفاعیلن مفاعیلن فعولن | ˘–––/˘–––/˘–– | 386 | 10.03 |
| *4* | مفعولُ مفاعیلُ مفاعیلُ فعولن | ––˘/˘––˘/˘––˘/˘–– | 334 | 8.68 |
| *5* | مفعولُ مفاعیلُ مفاعیلُ فعل | ––˘/˘––˘/˘––˘/˘– | 272 | 7.07 |
| *6* | مفعولُ فاعلاتُ مفاعیلُ فاعلن | ––˘/–˘–˘/˘––˘/–˘– | 213 | 5.54 |
| *7* | فعلاتن فعلاتن فعلاتن فعلن | ˘˘––/˘˘––/˘˘––/˘˘– | 138 | 3.59 |
| *8* | مفعولُ مفاعلن فعولن | ––˘/˘–˘–/˘–– | 131 | 3.41 |
| *9* | فاعلاتن فاعلاتن فاعلن | –˘––/–˘––/–˘– | 62 | 1.61 |
| *10* | فعلاتن مفاعلن فعلن | ˘˘––/˘–˘–/˘˘– | 45 | 1.17 |
| *11* | مفاعلن فعلاتن مفاعلن فعلن | ˘–˘–/˘˘––/˘–˘–/˘˘– | 40 | 1.04 |
| *12* | مفعولُ مفاعیلن مفعولُ مفاعیلن | ––˘/˘–––/––˘/˘––– | 31 | 0.81 |
| *13* | مفعولُ فاعلاتن مفعولُ فاعلاتن | ––˘/–˘––/––˘/–˘–– | 28 | 0.73 |
| *14* | فعلاتن فعلاتن فعلن | ˘˘––/˘˘––/˘˘– | 20 | 0.52 |
| *15* | مستفعلن مستفعلن مستفعلن مستفعلن | ––˘–/––˘–/––˘–/––˘– | 19 | 0.49 |
| *16* | فعولن فعولن فعولن فعل | ˘––/˘––/˘––/˘– | 14 | 0.36 |
| *17* | مفاعلن فعولن مفاعلن فعولن | ˘–˘–/˘––/˘–˘–/˘–– | 13 | 0.34 |
| *18* | مفتعلن فاعلن مفتعلن فاعلن | –˘˘–/–˘–/–˘˘–/–˘– | 9 | 0.23 |
| *19* | مفتعلن مفتعلن فاعلن | –˘˘–/–˘˘–/–˘– | 8 | 0.21 |
| *20* | فعولن فعولن فعولن فعولن | ˘––/˘––/˘––/˘–– | 8 | 0.21 |
| *21* | فاعلاتن فاعلاتن فاعلاتن فاعلاتن | –˘––/–˘––/–˘––/–˘–– | 7 | 0.18 |
| *22* | مفتعلن مفاعلن مفتعلن مفاعلن | –˘˘–/˘–˘–/–˘˘–/˘–˘– | 7 | 0.18 |
| *23* | مفاعلن مفاعلن مفاعلن مفاعلن | ˘–˘–/˘–˘–/˘–˘–/˘–˘– | 7 | 0.18 |
| *24* | مفاعیلُ مفاعیلُ مفاعیلُ فعولن | ˘––˘/˘––˘/˘––˘/˘–– | 5 | 0.13 |
| *25* | متفاعلن متفاعلن متفاعلن متفاعلن | ˘˘–˘–/˘˘–˘–/˘˘–˘–/˘˘–˘– | 3 | 0.08 |
| *26* | مفاعلن فعلاتن مفاعلن فعلاتن | ˘–˘–/˘˘––/˘–˘–/˘˘–– | 2 | 0.05 |
| *27* | فعلاتُ فاعلاتن فعلاتُ فاعلاتن | ˘˘–˘/–˘––/˘˘–˘/–˘–– | 2 | 0.05 |

**2-3 Meter Classification in Kurdish Poetry**

Experts of Persian poetry use the following traditional steps for identification of meter in quantitative verses (Elwell-Sutton, 1976):

1. Scansion: each line will be divided into its syllables.
2. Comparing: light and heavy syllables sequence is compared with the known common meter patterns.
3. Considering poetic license: Sometimes it is necessary to make changes in the pronunciation of certain words in order to match the overall meter of the poem such as making a light syllable heavy, lightening of a heavy syllable, and fade in together two adjacent words.

If a pattern is repeated in all lines, the poem will be recognized as a quantitative verse, and that pattern is proposed as the meter of the poem. Otherwise, if all lines have an equal number of syllables, then the poem will be recognized as a syllabic verse of that number. Else, when lines have neither a consistent pattern nor an equal number of syllables, then the poem is a free verse.

**2-4 Related Works**

As far as the authors know, no research has been done on the automatic classification of Kurdish poetry. Considering the similarities between the Kurdish quantitative verses and classic Persian, Arabic, and Turkish ones, we will give a brief overview of the works done in these languages.

The Arabic and Persian orthographies are abjad, and short vowels are written only for kids or ritual texts. Mojiri (2008), Kurt & Kara (2012), Alabbas et al. (2014) and Abuata & Al-Omari (2018) have consider preprocessing steps for insertion of short vowels and turning the text into phonemic representation. For example, The Basrah system (Alabbas et al., 2014) converts word “سدّ” to “سَدْد” and “والشمس” into “وَشْشمس”. Mojiri (2008) looks up the words that cannot be syllabified by the rules from a transliteration dictionary. Jafarari Qamsari (2015) relies on the distributive characteristics of Persian phonemes, and by using poetic and phonetic rules, converts the input Persian couplet into light, heavy, and potential heavy syllable string.

A key step in the meter classification is comparison with common patterns. Mojiri (2008) compares with 31 common Persian patterns, Alabbas et al. (2014) compares with 16 bahrs of Arabic, and (Kurt & Kara (2012) compares with 20 plain and 45 mixed Ottoman templates.

The efficiency of the works is varying. Mojiri (2008) reports 65% precision, Jafarari Qamsari (2015) accuracy of more than 98%, Alabbas et al. (2014) precision higher than 96%. Recently, data-driven and machine-learning works have been done on Arabic meter, e.g. Yousef et al. (2019) by Recurrent Neural Networks reports overall accuracy of 96.38% and Al-shaibani et al. (2020) by deep bidirectional recurrent neural networks reports more than 94% accuracy.

## 3 Kurdish Poem Meter classification

In this section, we describe the proposed method in detail. The input is a Kurdish poem text written in the standard alphabet of Kurdish. The output is the type (quantitative, syllabic, or free verse) and the metrical pattern of the poem. The traditional manual method described earlier influenced our method of automatic meter classification. Fig. 1 illustrates the flowchart of the proposed method. The method is available as a web application[1].

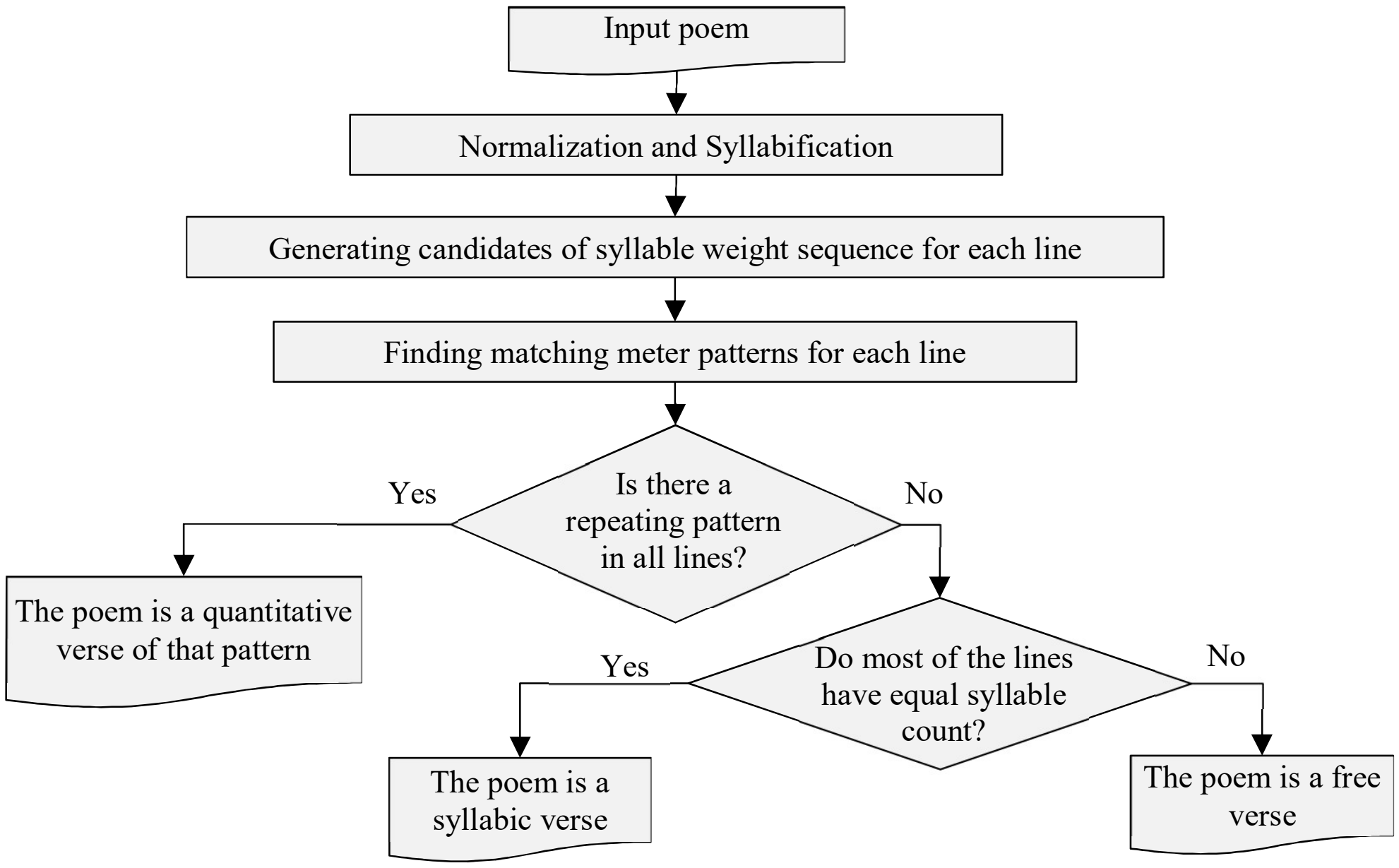

Fig. 1: Flowchart of the proposed method

### 3-1 Normalization and Syllabification

As the input of the proposed method is a plain text, we perform the following normalization steps for preventing errors:

- Removing lines that contain plenty of non-Kurdish characters or have previously tagged as non-Kurdish.
- Converting some Arabic specific characters to the standard Kurdish equivalents, e.g. “ث” and “ص” into ‘س’.
- Converting punctuations and end-of-line characters to a plus (open) juncture mark (+).

[1] https://asosoft.github.io/PoemClassifier/

Next is the syllabification process. For orthographies like English, Arabic, and Kurdish that do not have a one-to-one correspondence between the alphabet letters and the phonemes of the language, there will be challenges for the syllabification. In this study, we use the rule-based method of SCK grapheme-to-phoneme conversion presented by Mahmudi & Veisi (2021), which converts the input text into a syllabified string of phonemes. This method also correctly merges conjunction ‘و’ (and) with the previous word (Mahmudi & Veisi, 2021). This merge, e.g., “تیر و کەوان” */tî.rû ke.wan/* and “پۆڵاو ئاسن” */po.ław ʔa.sin/* is required for the following scansion step.

For example, the following line by Nalî (1800-1877) will be normalized and syllabified as:

the input: گەر نەبەخشێ مەرهەمی وەصڵی، برینم کارییە

normalized: گەر نەبەخشێ مەرهەمی وەسڵی+ برینم کارییە+

syllabified: *ger ne.bex.şê mer.he.mî wes.łî + bi.rî.nim ka.rî.ye +*

## 3-2 Generating Candidates of Syllable Weight Sequence

As vowel length is not distinctive in Kurdish, for preserving the meter in quantitative verses, sometimes, short vowels should be pronounced long and long ones short. There are some clues for recognizing syllable weight changes automatically:

A. Long vowels in word-final unstressed positions are pronounced short.
B. When a short vowel precedes a juncture (punctuations or the end of a line), it is usually pronounced long.
C. In quantitative verses that foot start with two adjacent light syllables, often the first syllable is heavy, and for satisfying the meter, it should be pronounced light.
D. When long vowel */î/* in an open syllable precedes approximant */y/*, the vowel is pronounced short. E.g., */nî.ye/* ‘is not’.
E. A syllable with two-consonant onset can also be pronounced as a two-syllable sequence (˘¯). For example, */xwa/* ‘god’ as */xu.wa/* and */gyan/* ‘soul’ as */gi.yan/*.

For managing the uncertainties in syllable weights, considering the above clues, we generate possible weight sequence candidates. For example, in */ger nebexşê merhemî wesłî birînim karîye/*, we have:

| | 1 | 2 | 3 | 4 | 5 | 6 | 7 | 8 | 9 | 10 | 11 | 12 | 13 | 14 | 15 |
|---|---|---|---|---|---|---|---|---|---|---|---|---|---|---|---|
| syllable | *ger* | *ne* | *bex* | *şê* | *mer* | *he* | *mî* | *wes* | *łî* | *bi* | *rî* | *nim* | *ka* | *rî* | *ye* |
| normal weight | ¯ | ˘ | ¯ | ¯ | ¯ | ˘ | ¯ | ¯ | ¯ | ˘ | ¯ | ¯ | ¯ | ˘ | ˘ |
| both weights are possible? | ✓ | | | | | | ✓ | | ✓ | | | | | ✓ | ✓ |
| meter pattern | ¯ | ˘ | ¯ | ¯ | ¯ | ˘ | ¯ | ¯ | ¯ | ˘ | ¯ | ¯ | ¯ | ˘ | ¯ |

- By clue C, syllable 1 can be pronounced light because we do not know the meter for now.
- By clue A, syllables 7 and 9 can be pronounced light.
- By clue D, syllable 14 is pronounced light.
- By clue B, syllable 15 can be pronounced heavy.

In the above example, for 5 syllables, there are 2 possible weights; therefore 2^5=32 sequence candidates can be generated.

## 3-3 Finding the Matching Patterns for Each Line

As the quantitative meters have more detailed and harder to compose, we first examine the lines of the poem for detecting a quantitative pattern. In this step, for each line, we compute Levenshtein edit distance of each syllables weight sequence candidate with 27 common meter patterns (presented in Table 5). For example, if a line of the poem has 32 weight sequence candidates, we must calculate 32×27=864 edit distances. Since the strings are less than 20 characters long and contain only two characters (˘ and ¯), these calculations are done quickly.

For each line, we only store candidate-pattern pairs that have the smallest distances bellow a maximum acceptable distance (given 4). For example, for */ger nebexşê merhemî wesłî birînim karîye/*, among 864 pairs, only 62 pairs are acceptable and one of them is:

- syllable weight sequence candidate: –ᴗ–––ᴗ–––ᴗ–––ᴗᴗ
- nearest common pattern: –ᴗ–––ᴗ–––ᴗ–––ᴗ– (with an edit distance of 1)

### 3-4 Meter Classification

The meter classification of a Kurdish poem, just by one or two lines is not correct at all the times, because:

- some lines of a syllabic verse may follow a quantitative pattern
- some lines may contain misspellings
- syllabification of some words and weight of some syllables are ambiguous
- unprofessional poets may commit mistakes in patterns

Therefore, in our proposed method, we consider all lines of the poem together. For each acceptable pair from the previous step, we add up to the score of the corresponding pattern for the whole poem. Eventually, there is a score for each common meter pattern. The pattern with the highest score is the most probable quantitative pattern of the poem; i.e., we define:

$$P = \underset{p_j \in M}{\arg\max} \sum_{i=1}^{n} \left(\text{MaxDist} - Dist(p_j, w_i)\right) \qquad (1)$$

In which, P is the most probable quantitative pattern of the poem, M is the set of common metrical patterns, n is the number of lines of the poem, Dist($p_j$, $h_i$) is the edit distance of a pattern ($p_j$) and weight sequence of a line ($w_i$), MaxDist is the maximum acceptable edit distances (given 4).

We calculate a regulated measure for

$$confidence = \frac{HighestScore}{MaxDist \times LinesCount} \qquad (2)$$

However, the poem must have the following conditions to be recognized as a “quantitative verse”:

- Nearly all lines of the poem must have a same syllable count, i.e., the amount of standard deviation has to be small.
- The majority of lines must comply with a pattern, i.e., the calculated confidence has to be high.

If a poem fulfills only the first condition, the proposed method assigns it as a “syllabic verse” of the statistical mode of syllables count of lines. Else, if none of the above conditions fulfill with the poem, it will be classified as a “free verse”.

## 4 Results

### 4-1 Test Dataset

We evaluated our proposed method on a dataset consisted of 1,154 Central Kurdish poems (979 quantitative, 130 syllabic, and 45 free verses) from available poems of “VejinBooks.com”. This website is a growing free online corpus of Kurdish literary texts in different dialects of Kurdish. The type and meter of all poems in this corpus are specified manually. VejinBooks also has statistics about the frequency of each meter available at the corpus. Among the available texts inside the website, we chose only poems with more than three couplets from 12 well-known poets of Central Kurdish. Table 6 shows the overall statistics of the dataset. The dataset is available in Github[2].

[2] https://github.com/AsoSoft/Vejinbooks-Poem-Dataset

Table 6: Statistics of poems of the dataset

| poet | quantitative | syllabic | free verse |
|---|---|---|---|
| Nalî (1800-1877) | 125 | - | - |
| Salim (1800-1866) | 259 | - | - |
| Kurdî (1809-1850) | 77 | - | - |
| Ḧacî Qadir (1816-1897) | 102 | - | - |
| Wefayî (1844-1902) | 118 | 12 | - |
| Ḧerîq (1856-1909) | 49 | 1 | - |
| Narî (1874-1944) | 95 | - | - |
| Qaniɛ (1898-1965) | 56 | 13 | - |
| Diłdar (1918-1948) | 18 | 11 | - |
| Hêmin (1921-1986) | 53 | 29 | - |
| Herdî (1922-2006) | 13 | - | - |
| Kakey Felaḧ (1928-1990) | 14 | 64 | 45 |
| ***Overall*** | ***979*** | ***130*** | ***45*** |

The use of Arabic or Persian phrases (like Arabic Quranic Verses) within the text is a common convention in Kurdish poetry. Since our method is based on Central Kurdish phonology, this can be a problem for the evaluation. Fortunately, in VejinBooks corpus, non-Kurdish phrases are tagged. We removed all the lines that had a non-Kurdish phrase inside the test dataset.

**4-2 Test Results**

We evaluated our method in type (quantitative, syllabic, or free) and pattern classification. The evaluation metrics are precision, recall, and F1-score. In Table 7, we show the results of the poem type classification.

Table 7: Test Results for poem type classification

| poem type | count | precision (%) | recall (%) | F1-score (%) |
|---|---|---|---|---|
| quantitative | 979 | 99.4 | 97.4 | 98.4 |
| syllabic | 130 | 83.2 | 95.4 | 88.9 |
| free verse | 45 | 100 | 100 | 100 |
| ***overall*** | ***1,154*** | ***97.3*** | ***97.3*** | ***97.3*** |

Table 8 indicates the test results for pattern classification. It shows the efficiency of the proposed method for each pattern. The recall for patterns that have “فعلاتن” (˘˘¯¯) or “فعلن” (˘˘¯) feet, like “ فعلاتن مفاعلن فعلن”, is low. The method often classifies the poems of these patterns as syllabic. It causes the lower precision for syllabic type classification, as shown in Table 7. Maybe the reason is finding and matching words in the poem with two adjacent light syllables in the start of feet is hard in Kurdish. Therefore, poets consider using poetic licenses to preserve the meter.

Table 8: Meter pattern classification results, separated by pattern

| **meter pattern** | **count** # | **precision** % | **recall** % | **F1-score** % |
|---|---|---|---|---|
| مفاعيلن مفاعيلن مفاعيلن مفاعيلن | 245 | 100 | 100 | 100 |
| فاعلاتن فاعلاتن فاعلاتن فاعلن | 224 | 99 | 100 | 99 |
| مفاعيلن مفاعيلن فعولن | 151 | 99 | 100 | 99 |
| مفعولُ مفاعيلُ مفاعيلُ فعولن | 117 | 99 | 99 | 99 |
| فعلاتن فعلاتن فعلاتن فعلن | 77 | 100 | 91 | 95 |
| فاعلاتن فاعلاتن فاعلن | 33 | 92 | 100 | 96 |
| مفعولُ فاعلاتُ مفاعيلُ فاعلن | 31 | 100 | 100 | 100 |
| فعلاتن مفاعلن فعلن | 25 | 100 | 20 | 33 |
| مفعولُ مفاعيلن مفعولُ مفاعيلن | 16 | 100 | 100 | 100 |
| فعلاتن فعلاتن فعلن | 13 | 100 | 77 | 87 |
| مفعولُ فاعلاتن مفعولُ فاعلاتن | 10 | 90 | 90 | 90 |
| مفاعلن فعلاتن مفاعلن فعلن | 7 | 100 | 57 | 73 |
| مفعولُ مفاعلن فعولن | 5 | 100 | 100 | 100 |
| مفتعلن مفاعلن مفتعلن مفاعلن | 4 | 100 | 75 | 86 |
| مفتعلن فاعلن مفتعلن فاعلن | 4 | 80 | 100 | 89 |
| مستفعلن مستفعلن مستفعلن مستفعلن | 4 | 80 | 100 | 89 |
| فعولن فعولن فعولن فعل | 3 | 75 | 100 | 86 |
| فاعلاتن فاعلاتن فاعلاتن فاعلاتن | 3 | 100 | 100 | 100 |
| مفتعلن مفتعلن فاعلن | 2 | 100 | 100 | 100 |
| فعلاتُ فاعلاتن فعلاتُ فاعلاتن | 2 | 100 | 50 | 67 |
| مفعولُ مفاعيلُ مفاعيلُ فعل | 1 | 100 | 100 | 100 |
| مفاعيلُ مفاعيلُ مفاعيلُ فعولن | 1 | 0 | 0 | 0 |
| مفاعلن فعولن مفاعلن فعولن | 1 | 25 | 100 | 40 |
| مفاعلن مفاعلن مفاعلن مفاعلن | 0 | 0 | 0 | 0 |
| مفاعلن فعلاتن مفاعلن فعلاتن | 0 | 0 | 0 | 0 |
| 16-syllabic | 9 | 100 | 89 | 94 |
| 15-syllabic | 6 | 100 | 100 | 100 |
| 14-syllabic | 8 | 55 | 75 | 63 |
| 13-syllabic | 4 | 100 | 100 | 100 |
| 12-syllabic | 5 | 100 | 100 | 100 |
| 11-syllabic | 8 | 100 | 63 | 77 |
| 10-syllabic | 49 | 71 | 100 | 83 |
| 8-syllabic | 31 | 100 | 100 | 100 |
| 7-syllabic | 10 | 100 | 100 | 100 |
| free verse | 45 | 100 | 100 | 100 |
| ***overall*** | | ***96.2*** | ***96.2*** | ***96.2*** |

Table 9 shows the test results for metrical pattern classification, separated by authors. It can be speculated that how much a poet complies with the patterns and uses fewer poetic licenses. For example, Herdî is known for having few but admirable poems. The lower accuracy for the poems of Hêmin and Ḧacî Qadir is duo to using patterns with “فعلاتن” (˘˘¯¯) or “فعلن” (˘˘¯) feet and using more poetic licenses.

Table 9: Metrical pattern classification results, separated by poet

| Author | F1-Score |
|---|---|
| Diłdar | *100* |
| Herdî | *100* |
| Nalî | *99.2* |
| Kakey Felaḧ | *99.2* |
| Kurdî | *98.7* |
| Ḧerîq | *98.0* |
| Salim | *97.7* |
| Qaniɛ | *97.1* |
| Narî | *96.8* |
| Wefayî | *96.2* |
| Ḧacî Qadir | *87.3* |
| Hêmin | *86.6* |
| ***Overall*** | ***96.2*** |

## 5 Conclusions and Future Works

In this paper, we have proposed an automatic poem meter classifier for the Central Kurdish language. The evaluations achieved an overall precision of 97.3% in meter type classification and overall precision of 96.2% in metrical pattern identification. Now, this algorithm assists the contributors of Vejinbooks online corpus in tagging newly imported poems. Furthermore, an online application is developed for amateur poets for evaluating their poems.

In the future, we plan to extend the functionality of the method in identifying subclasses of Kurdish free verses. Automatic author identification, based on characteristics of the poem is another field of study for further works.

## Acknowledgements

We thank contributors of “Vejîn Culture and Arts Institute” for providing manually-checked dataset for evaluating our proposed method.

## Data Availability Statement

The data that support the findings of this study are openly available in “GitHub” at `https://doi.org/10.5281/zenodo.4079471`, reference number 4079471.